\documentclass[sigconf,nonacm]{acmart}

\acmConference{}{}{}
\acmBooktitle{}
\acmPrice{}
\acmISBN{}
\acmDOI{}

\AtBeginDocument{%
  }

\usepackage{amsmath,amssymb}
\usepackage{tikz}
\usepackage{pgfplots}
\pgfplotsset{compat=1.18}

\begin{document}

\title{E-CHUM: Event-based Cameras for Human
Detection and Urban Monitoring}

\author{Jack Brady}
\affiliation{%
  \institution{University of Michigan}
  \city{Ann Arbor}
  \state{MI}
  \country{USA}
}

\author{Andrew Dailey}
\affiliation{%
 \institution{University of Michigan}
 \city{Ann Arbor}
 \state{MI}
 \country{USA}}

\author{Kristen Schang}
\affiliation{%
  \institution{University of Michigan}
  \email{keschang@umich.edu}
  \city{Ann Arbor}
  \state{MI}
 \country{USA}}

\author{Zo Vic Shong}
\affiliation{%
  \institution{University of Michigan}
 \email{zovic@umich.edu}
  \city{Ann Arbor}
  \state{MI}
 \country{USA}}


\begin{abstract}
 Understanding human movement and city dynamics has always been challenging. From traditional methods of manually observing the city's inhabitant, to using cameras, to now using sensors and more complex technology, the field of urban monitoring has evolved greatly. Still, there are more that can be done to unlock better practices for understanding city dynamics.

 This paper surveys how the landscape of urban dynamics studying has evolved with a particular focus on event-based cameras. Event-based cameras capture changes in light intensity instead of the RGB values that traditional cameras do. They offer unique abilities, like the ability to work in low-light, that can make them advantageous compared to other sensors. 

Through an analysis of event-based cameras, their applications, their advantages and challenges, and machine learning applications, we propose event-based cameras as a medium for capturing information to study urban dynamics. They offer the ability to capture important information while maintaining privacy.

We also suggest multi-sensor fusion of event-based cameras and other sensors in the study of urban dynamics. Combining event-based cameras and infrared, event-LiDAR, or vibration has to potential to enhance the ability of event-based cameras and overcome the challenges that event-based cameras have.

\end{abstract}



\maketitle

\section{Introduction}
 Understanding human movement and city dynamics has always been challenging. Traditional urban studies depended on people standing in public spaces, watching crowds, taking notes, or setting up cameras to capture behavior. These approaches produced valuable insights but required significant time and effort, and they often captured only brief moments rather than continuous patterns.

Today, cities can be monitored using sensors and data-driven tools that operate with far less manual work. Common methods include video cameras, environmental sensors like air-quality monitors, and location data from mobile devices. Each method has strengths but also limitations. Standard cameras offer detailed imagery but struggle in low light or glare and generate large amounts of data that raise storage and privacy concerns. Mobile or GPS data can reveal how people move through a city, but collecting it requires consent and raises ethical issues primarily regarding privacy. Even indirect sensing techniques that infer human presence from environmental signals can be inconsistent or intrusive.

Overall, many conventional sensing tools force difficult trade-offs among data richness, lighting conditions, and personal privacy.

This survey looks at a newer technology that may help address these challenges: event-based cameras. We explain how event cameras work, examine their advantages and limitations for urban monitoring, and describe how machine learning can convert event streams into meaningful information about people and activity patterns. We close by highlighting emerging possibilities, including combining event cameras with other sensing methods to unlock more capable and privacy-aware tools for understanding city dynamics.

\section{Previous Work on Urban Dynamics}
The study of urban dynamics is not a new research field. However, it has evolved significantly as the technology to monitor city spaces has developed. The first studies of urban spaces was done through manual observation. This was a demanding process which required large amounts manpower. One example of this was Stanley Milgram's research of New York in 1977 \cite{Milgram1977}. William Whyte used cameras for his urban studies research in 1980. He had set up cameras and set them to record time-lapses \cite{Whyte1980}. The trouble with both of these examples' methods is that they require lots of time and effort to sift through the data after it has been collected. The effort aside, it is difficult to find trends or observations that one is not explicitly looking for. If looking to confirm a hypothesis, the researcher may miss a byproduct result because they are not looking for it in their manual analysis of the data. Additionally, there are privacy concerns. William Whyte's work captured lots of video data of the people he was studying. While he was in public areas, it would have been unfeasible to obtain consent from all that he video-graphed.

With the advancement of technology, other ways of sensing has been used to monitor and learn about city dynamics. The widespread use of mobile devices has allowed for the ability to capture the locations of people simply by monitoring the GPS data on things like their cell phones. The Livehoods Project \cite{Cranshaw_Schwartz_Hong_Sadeh_2021} did exactly this. They used social media to capture the location data of people in Pittsburgh. They were able to group people in clusters based off of what boundaries people did or didn't cross. They were able to draw neighborhoods by the behavior of the people as opposed to municipal boundaries. Another recent work monitored air quality to detect homeless populations in cities \cite{gersey2025surveycitywidehomelessnessdetection}. The drawbacks of these more modern approaches is the scope in which the collected data. The Livehoods Project was only able to gather data on those with cell phones and willingly participated in the research. As their research describes, there are missing neighborhoods where users had a lower use of cell phones. The research with air quality measures also was limited in scope in that it could only detect a specific circumstance and would struggle with changes in the environment, as described in their paper as well.

Another recent work was trash detection in cities using computer vision to explore city sanitation \cite{10.1145/3643832.3661431}. This work focused on the ability to automate the labeling process of objects that are detected through computer vision algorithms. The labeling process even graded frames which allows the researcher to sift through data efficiently and understand areas of importance. The automated labeling and grading provided a way to sort through large amounts of data. In this paper, the dataset used was from videos of regular cameras mounted to cars. A benefit of this is that data capture doesn't impact the environment, but there could be concerns about privacy.  

Prior work varies in scope, method of data collection, and flaws. As a whole, as the scope grows, so do concerns over privacy, but the biggest challenge with a large scope is the sheer amount of data that is captured, which is inherent with images and other high fidelity data collection types. As the scope shrinks, so do the concerns that the large scope methods hold, but new concerns arise. For example, a sensor that can only captures air quality would miss understanding people's posture. In the end, prior work laid the foundation of city dynamics research and offered many new insights into urban environments, but are limited by the technology that they employ.

\section{Introduction to Event-Based Cameras}
\subsection{Operation Principles}
\begin{figure}
    \centering 
    \includegraphics[width=.9\linewidth]
    {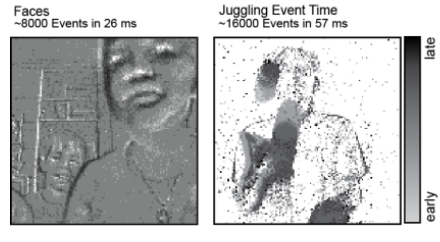}
    \caption{Example of Event Frames \cite{10.1007/978-3-642-33863-2_52}}
    \label{fig:placeholder}
\end{figure}

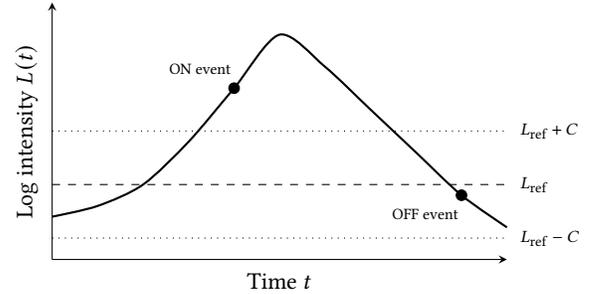
\begin{figure}[t]
    \centering
    \begin{tikzpicture}
        \begin{axis}[
            width=0.9\linewidth,
            height=5cm,
            xlabel={Time $t$},
            ylabel={Log intensity $L(t)$},
            xmin=0, xmax=10,
            ymin=-0.2, ymax=2.2,
            axis lines=left,
            xtick=\empty,
            ytick=\empty,
            clip=false
        ]

        \addplot[smooth,thick]
            coordinates {
                (0,0.2) (1,0.3) (2,0.5) (3,0.9)
                (4,1.4) (5,1.9) (6,1.6) (7,1.2)
                (8,0.8) (9,0.4) (10,0.1)
            };

        \addplot[dashed]
            coordinates {(0,0.5) (10,0.5)};
        \node[anchor=west, xshift=2pt] at (axis cs:10,0.5)
            {\scriptsize $L_{\text{ref}}$};

        \addplot[dotted]
            coordinates {(0,1.0) (10,1.0)};
        \node[anchor=west, xshift=2pt] at (axis cs:10,1.0)
            {\scriptsize $L_{\text{ref}} + C$};

        \addplot[dotted]
            coordinates {(0,0.0) (10,0.0)};
        \node[anchor=west, xshift=2pt] at (axis cs:10,0.0)
            {\scriptsize $L_{\text{ref}} - C$};

        \addplot[mark=*,only marks]
            coordinates {(4,1.4)}
            node[above left, yshift=2pt, xshift=2pt]
            {\scriptsize ON event};

        \addplot[mark=*,only marks]
            coordinates {(9,0.4)}
            node[below left, yshift=-2pt, xshift=2pt]
            {\scriptsize OFF event};

        \end{axis}
    \end{tikzpicture}
    \caption{Illustration of the event-generation principle for a single pixel.
    The pixel maintains a stored reference log-intensity $L_{\text{ref}}$ and
    triggers an ON (brightness increase) or OFF (brightness decrease) event
    whenever the current log intensity $L(t)$ deviates from $L_{\text{ref}}$
    by more than a positive or negative contrast threshold $C$.}
    \label{fig:log-intensity}
\end{figure}

Event-based cameras differ from conventional frame-based sensors in that they do not acquire full images at a fixed rate. Instead, each pixel independently monitors changes in brightness over time and reports only those changes that are sufficiently large. An example of frames from event cameras can be seen in Figure ~\ref{fig:placeholder}. When the log intensity value at a pixel deviates from its last stored value by some threshold, the pixel will emit its information (a digital “event”) out to the shared bus. As one might expect, these outputs aren’t uniformly distributed, with the data rate driven by scene dynamics rather than an external clock signal \cite{Gallego_2022, Chakravarthi_2024}. Figure~\ref{fig:log-intensity} illustrates this process for a single pixel: the log-intensity signal evolves over time, and events are triggered whenever it crosses positive or negative contrast thresholds relative to the last event value.

At the circuit level, incoming photons are converted to current in a photodiode and then to a voltage signal. This voltage is processed in the logarithmic intensity domain and continuously compared to a reference corresponding to the log-intensity value at the time of the last event. Each pixel checks if the difference between the current log intensity L and the stored reference $L_{\text{ref}}$ exceeds a positive or negative threshold C. If it does, the pixel triggers an ON (brightness increase) or OFF (brightness decrease) event and then later updates Lref for the next brightness reading \cite{Gallego_2022, Chakravarthi_2024}.

Every pixel on the camera outputs an event tuple
$e_k = (x_k, y_k, t_k, p_k)$, where $(x_k, y_k)$ are the pixel coordinates,
$t_k$ is the timestamp (typically with microsecond resolution), and
$p_k \in \{-1,+1\}$ encodes the polarity of the brightness change.
Captured events are transmitted from the 2D array of pixels to peripheral
circuitry at the edge of the chip and off-chip through a shared digital bus,
commonly using an address–event representation (AER) protocol~\cite{Gallego_2022}.
Because the sensor is data-driven, the aggregate event rate varies with the
amount of motion or brightness change in the scene and can range from a few~MHz
to the~GHz range, depending on the hardware and interface~\cite{Gallego_2022}.

The irradiance reaching each pixel can be factored into a slowly varying illumination term and a spatially varying surface reflectance term. Under the common assumption that illumination is approximately constant over short time intervals, changes in log intensity are dominated by changes in reflectance due to object or edge motion within the field of view \cite{Gallego_2022}. As with any physical photoreceptor, however, the analog front-end has finite temporal bandwidth. When the rate of each event fluctuates faster than the permitted bandwidth of the circuit, the photoreceptor will run the signal through a low-pass filter. As the modulation frequency increases, approaching the cutoff frequency, the resulting event rate per cycle drops \cite{Gallego_2022}.

Unlike frame-based cameras, which sample the scene at discrete exposure times and therefore exhibit temporal aliasing for content above the Nyquist frequency, event cameras operate in effectively continuous time: events are generated whenever the threshold condition is met, without waiting for a global exposure. They still ignore variations beyond their analog bandwidth, but those high-frequency components are attenuated rather than folded back into lower frequencies \cite{Gallego_2022}. 

\subsection{Applications of Event-Based Cameras}
Event-based cameras play concrete roles in robotics, automotive perception, 3D vision, and high-speed scientific imaging. Because of their microsecond temporal resolution and high dynamic range, recent survey work on event-based vision and datasets has found that they are particularly attractive for situations where objects are moving fast, challenging to illuminate, or limited on power and bandwidth \cite{Gallego_2022, Chakravarthi_2024}.
	
As mentioned earlier, an important application area is robot perception, especially visual odometry and SLAM. Event cameras are generally used either alone or in conjunction with grayscale frames and inertial sensors; the goal of these sensors is to estimate motion in real time, often at faster speeds and lower latencies than frame-based systems can handle \cite{Gallego_2022, Chakravarthi_2024}. Public datasets that provide asynchronous events, global-shutter frames, IMU readings, and ground-truth poses for a variety of indoor and outdoor trajectories have become standard benchmarks for event-based pose estimation, visual odometry, and SLAM \cite{Mueggler_2017}. More recent work has explored the use of event-camera odometry for legged and wheeled robots in planetary exploration, highlighting the usefulness of this sensor under highly dynamic motion and extreme lighting conditions \cite{Mahlknecht_2022}. 

In the automotive industry, event cameras are being used as complementary sensors for driver-assistance and autonomous driving. Survey work in this domain has highlighted how event-based cameras are used for tasks such as forward perception, object detection, driver monitoring, and fusion with lidar and radar. \cite{Shariff_2023} Large real-world datasets based on commercial event sensors further illustrate that event cameras are now being tested in realistic on-road scenarios rather than only in controlled laboratory settings; these datasets indicate these sensors were used for traffic monitoring, detection of vulnerable road users, and nighttime driving \cite{Chakravarthi_2024}.

Another critical application area for event cameras is depth estimation and 3D perception. Reviews of event-camera-based depth estimation methods describe stereo, monocular, and multi-sensor approaches that exploit the fine timing of events to recover depth; these methods often aim to solve joint motion-and-structure problems or combine events with frames and lidar \cite{Hong_2022}. Datasets such as MVSEC and other event-based stereo benchmarks demonstrate how these methods can provide dense depth maps and 3D reconstructions in scenes that are difficult for stereo matching; these situations include low-light and high-dynamic-range environments \cite{Chakravarthi_2024}.

Beyond vehicle navigation and 3D geometry, event cameras are also used in specialized scientific and individual settings, where precise timing is crucial. One example uses an event-based sensor for velocity-resolved kinetics measurements. They elected to use these sensors because they could use their asynchronous stream to resolve reaction dynamics that evolve too quickly for conventional imaging \cite{Golibrzuch_2019}. Event-based vision has also been explored for tasks such as high-speed inspection, motion capture, gesture and action recognition, lip-reading, and object classification. All of these tasks are supported by a growing ecosystem of real-world and synthetic datasets covering human motion, traffic scenes, and standard vision benchmarks converted to event format \cite{Gallego_2022, Chakravarthi_2024}.

\subsection{Advantages of Event-Based Cameras}
Event-based cameras are unique in that they respond asynchronously to scene changes rather than reading out their values over fixed intervals. As a result, the design offers many advantages, including superfine temporal resolution, very low latency, high dynamic range, and low power consumption. Another important note is that these sensors can maintain sparse, adaptive data streams, which does not apply to frame-based cameras \cite{Gallego_2022, Chakravarthi_2024}.

The first advantage of these devices is their high temporal resolution with low latency. Unlike frame-based cameras, which integrate over a frame period, event cameras timestamp activities with microsecond precision \cite{Gallego_2022}. In fact, there are reports of latencies in the 10 µs to sub-millisecond range in systems today \cite{Gallego_2022}. Because events highlight small incremental changes rather than long-exposure averages, motion blur is largely avoided even at very high apparent motion \cite{Gallego_2022, Chakravarthi_2024}. From a systems standpoint, this means that both the temporal sampling of the scene and the responsiveness of the output stream are governed by scene dynamics rather than by an externally imposed frame rate.

A second significant benefit is low power consumption. Since no data is generated for static regions, the sensor’s dynamic power scales primarily with the amount of change in the scene, and static scenes yield negligible event rates. Gallego et al. reported chip-level power at approximately 10~mW for typical event-camera dies. For certain designs operating under specific regimes, they were able to achieve values below 10~µW \cite {Gallego_2022}. Embedded systems that integrate an event sensor with a lightweight processor can operate with total power budgets around 100~mW \cite{Gallego_2022}. This is substantially lower than many high-speed frame-based pipelines, which incur a continuous cost to acquire, transmit, and process dense images.

Unlike frame-based cameras, the event camera output makes it better suited for situations with limited bandwidth and storage. Because events are transmitted only when brightness changes exceed a threshold, the data rate is a proxy for scene activity. When the environment is static, the number of output events is low. When edges move, the output events are large and dense \cite{Gallego_2022, Chakravarthi_2024}. As a result, event streams typically contain far fewer samples than would be needed from a conventional camera to achieve comparable temporal fidelity. This property makes event cameras particularly suitable for platforms with limited storage and bandwidth.

Event-based pixels also generally have a high dynamic range. As mentioned earlier, these pixels respond quickly to changes in intensity and also work in the logarithmic domain. Thus, they can encode contrast over illumination ranges spanning several orders of magnitude. Typical DVS-like sensors achieve dynamic ranges greater than 120 dB, while high-quality frame-based cameras have ranges of 60-90dB \cite{Gallego_2022, Chakravarthi_2024}. Event cameras have no single global exposure and therefore the saturation and underexposure for each pixel are significantly reduced.

Another notable quality of event-based cameras is their robustness in challenging environments. Conventional cameras generally struggle to produce high-quality images in low light or with fast-moving objects. Because event-based cameras have lower latency and higher dynamic range, they produce a much more usable signal \cite{Gallego_2022, Chakravarthi_2024}. While these event cameras have their issues, such as quantization and bandwidth limitations, their modality is inherently well-suited to representing high-speed and high-contrast structure in the visual scene.

Finally, event cameras can offer potential privacy advantages. Their output consists of sparse streams of motion-related events rather than full-color, texture-rich images produced by conventional images. This would suggest that the observable details of scene appearance and identity are generally more concealed in event camera data than in conventional RGB video \cite{Gallego_2022, Chakravarthi_2024}. At the same time, however, recent work on reconstructing intensity images from events shows that appearance can sometimes be recovered to a surprising degree \cite{Rebecq_2019, 8946715, Scheerlinck_2020}. Event cameras should be considered privacy-enabling rather than intrinsically privacy-preserving: by default they reduce the amount of directly interpretable appearance information, but they do not provide a formal privacy guarantee. In practice, the level of protection they afford depends on system-level design choices, including which event data are stored, how long they are retained, where processing occurs (on-device vs. in the cloud), and what algorithms are allowed to run on the data.

\subsection{Challenges of Event-Based Cameras}
While event-based cameras offer a wide variety of advantages, they also host several challenges. The first is their unique output: an asynchronous stream of events in the form of tuples. As mentioned by Gaelgo et al. and Chakravarthi, the methods used to analyze information from classical, frame-centric vision systems cannot be applied to these event-based systems. End users must use or develop algorithms that preserve the event streams’ spatiotemporal structure \cite{Gallego_2022, Chakravarthi_2024}. There are a handful of workarounds for this problem, such as accumulating events into voxel grids or event frames. Unfortunately, all of these techniques reintroduce discretization in time and space, which ultimately blurs the advantages of the native event representation \cite{Gallego_2022}.

A second difficulty is photometric interpretation. Events encode threshold crossings of log intensity, so the observed stream depends jointly on local contrast, absolute brightness, and the relative motion between the sensor and the scene \cite{Gallego_2022}. The same physical structure can therefore generate very different event patterns under different motion or texture conditions. This makes tasks such as reconstructing absolute brightness or performing classical irradiance-based reasoning significantly more complex than in frame-based imaging. Many approaches rely on temporal accumulation, explicit generative models of event formation, or learning-based reconstructions, each of which introduces additional assumptions and computational overhead \cite{Gallego_2022, Rebecq_2019}.

Another issue with event cameras is the noise introduced by photons and sensor non-idealities. Additionally, the process of quantizing temporal contrast to discrete events has several issues, including pixel-to-pixel threshold mismatch, temporal jitter, and background activity \cite{Gallego_2022}. Similarly, when the event rate is high, the address-event buffers can saturate the data, leading to dropped events and timing distortions \cite{Gallego_2022}. Consequently, these issues have led to work in event denoising, robust event representations, and noise-characterization benchmarks. Unfortunately, standard models and calibration procedures are still less developed than those for conventional frame-based cameras \cite{Chakravarthi_2024}.

In addition to the sensor’s limitations, there are also issues with the event cameras ecosystem and prominent learning challenges. Datasets for frame-based cameras are plentiful and mainstream, while datasets with labeled event information are relatively small and task-specific. Naturally, the next course of action would be to turn to event-based simulators, which generate synthetic event-camera data. Unfortunately, event simulators still show noticeable gaps in their outputs relative to real sensor behavior \cite{Gallego_2022, Chakravarthi_2024, Rebecq_2018}. Most deep neural network models today are designed for dense images. Altering these architectures to be compatible with event streams requires non-standard encodings as well as careful choices about temporal aggregation \cite{Zheng_2023}. Finally, raw event data is too complex for humans to interpret directly. For users to make sense of this information, they look to reconstruction and visualization tools, such as E2VID and Prophesee’s Metavision Studio \cite{Rebecq_2019, PropheseeMetavisionStudio}. Similar to the deep architectures, these tools, while helpful, are significantly less developed than those used with conventional cameras \cite{Chakravarthi_2024}. In conclusion, event cameras have many advantages, however, they will require more specialized algorithms and software infrastructure to be a reliable alternative to frame-based cameras.

\section{Advantages of Event-Based Cameras for Urban Dynamics Research}
As previously shown, prior projects that are able to cover large areas are generally limited in the type of data that they can collect, relying on users to opt into using certain apps or reporting their data \cite{Cranshaw_Schwartz_Hong_Sadeh_2021, gersey2025surveycitywidehomelessnessdetection}. Projects of such scope do not struggle with as much of a privacy concern, since all data collection is done directly with the express consent of the parties reporting or opting into data collection. Other projects limit their data collection to data with fewer privacy concerns, such as vibrational data \cite{10.1145/3600100.3623750}. Such prior work, however, is limited in what it can accomplish. Other approaches to city monitoring systems require significant datasets free from the sample bias of opt-in systems. A common approach to mitigate such issues is the use of widespread cameras, taking photographs or video of areas to be processed \cite{Zhang27052024, inbook-duarte}. This technology is often paired with GPS and machine learning to build a comprehensive state that can be later analyzed \cite{Zhang27052024}. However, widespread projects utilizing camera vision often experience resistance due to concerns over the lack of privacy that is inherent in public photography. Deployments of city monitoring projects may require significant scope in the areas that are surveyed, which can lead to privacy concerns among surveilled populations \cite{kitchin2016ethics, KASHEF2021106923, IVANOVSKA2025103586}.

\begin{figure}[t]
  \centering
  \includegraphics[width=0.75\linewidth]{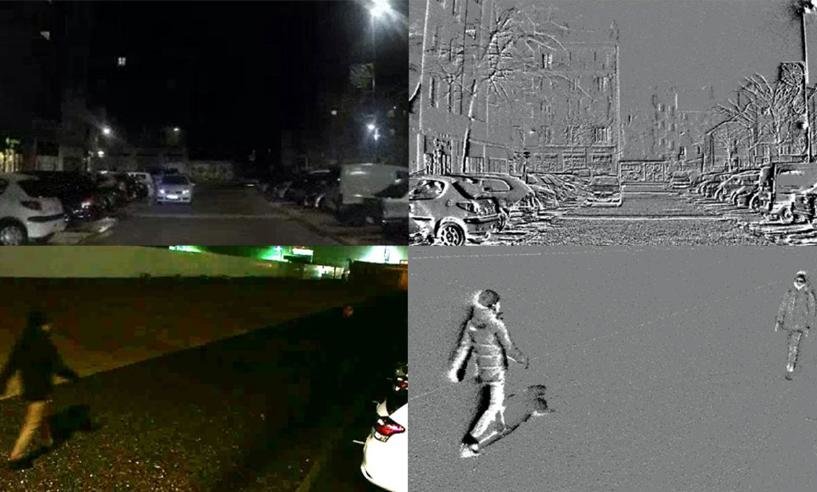}
  \caption{Difference shown between a traditional RGB image and event-based camera. Only a change in scenery is shown, obfuscating some identifiable data. Image courtesy of Prophesee France~\cite{eventcam_figure_researchgate}.}
  \Description{Side-by-side comparison of images from RGB and event-based cameras.}
  \label{fig:event_vs_rgb}
\end{figure}

Generally, the concerns over such city-wide cameras relate to privacy and monitoring of individuals. When equipped with facial recognition technologies, widespread cameras inherently allow for tracking of an individual throughout a city \cite{KASHEF2021106923}. Thus, any deployment of widespread camera nodes requires significant cybersecurity and privacy overhead to ensure the data is kept private, if allowed at all. Other approaches to privacy in surveillance generally include blurring, obfuscating, or removing identity-driving characteristics from a video stream \cite{PADILLALOPEZ20154177, fitwi, DelussuPutzuFumera2024SyntheticDataVideoSurveillance}. These methods may dilute data or may cause reduced quality in a model's results. Such methods also require secure data handling prior to this masking and often include storage of raw video footage, even if temporarily. Instead, event-based cameras may provide a means to avoid such problems from the data collection stage, as they capture less identifiable data in a single frame \cite{kim2025privacypreservingvisuallocalizationevent}. This can be seen illustrated in Figure~\ref{fig:event_vs_rgb}. Without the ability to identify an individual, which is generally done via facial recognition, the data coming from event-based cameras is decidedly more secure \cite{KASHEF2021106923, fitwi}. Modern work has not shown the ability for event-based cameras to generate facial recognition data, although research is ongoing and promising for face and eye pose detection \cite{s24051409}. Some lingering privacy concerns over event based cameras also include possible video reconstruction techniques, which may allow videos to be regenerated from event camera footage \cite{ZhangWangYangShenWen2023EVPerturb}. Still, footage from event cameras contains less identifiable data, which has not yet been proven to be able to identify individuals, meaning that such footage is deemed safer for widespread use, such as that in city-wide environments.

Urban dynamics research may also be done as a mobile field in dynamic environments. Prior work using traditional cameras can struggle with low light and changing light environments, as well as change of scenery \cite{Zhang_2017, DIMAURO2020175}. City-based environments are often monitored via mobile deployments, which must deal with scenery changes throughout diverse areas of a city, as well as changing light conditions both in time of day and area \cite{chen2019asc}. While event based cameras often struggle in moving environments, they are known to operate well in low light environments \cite{sun2025lowlightimageenhancementusing, Liu2024SeeingMotionAtNighttime}. They also are able to isolate dynamic environments by only capturing the moving part of an image and are known to eliminate motion blur that may be caused by a mobile deployment \cite{sundar2024generalizedeventcameras}.

Mobile, temporary, and other deployments not connected to the grid also must deal with energy as a primary resource, as traditional cameras are generally power intensive \cite{sanmiguel, 10.1145/990064.990096, izuka}. Event-based cameras can operate on lower power than traditional cameras, and as such provide benefits to deployments that operate on energy scavenging or battery power \cite{Gallego_2022, kim2025privacypreservingvisuallocalizationevent}.

\section{Machine Learning for Event-Based Cameras}
In addition to event-based cameras being a useful tool, machine learning is a useful tool for digesting the data from the event-based cameras. For example, computer vision is a useful machine learning method for digesting photo frames. Event-based cameras are highly incompatible with models made for traditional cameras because event-based cameras are asynchronous \cite{10517639}. This means that machine learning algorithms that take in traditional camera frames are incompatible to the frames from event cameras. Therefore, the machine learning strategies that have been used in traditional cameras have to be re-applied to event-based cameras and they largely have. 

Much effort has been applied to determining the optical flow of frames from event-based cameras. Optical flow is used to determine the motion of the objects captured and there have been a variety of strategies to understanding this feature in event-based camera frames \cite{Stoffregen2018, GUAMANRIVERA2025129899, 10517639, DBLP:journals/corr/abs-1802-06898, 8660483}. This work can be adapted to understanding city dynamics because understanding the motion of the frames captured could allow researchers to use event-based cameras to understand the movement of people and crowds within cities.

In addition, object detection and object tracking algorithms have been adapted to event-based camera frames. Even more specific, these algorithms have been applied to scenarios when the camera is moving or when the object detected is moving \cite{8593805, e-TLD, 10.1007/978-981-13-6553-9_18, 9607843}. For understanding urban dynamics, the events collected in the images that event-based cameras output can allow researchers to easily search for specific objects. Or it can allow researchers to detect people in images or count the number of people. In the end, the object detection that can observed from event-based cameras is useful for researchers to draw conclusions and understand patterns in city dynamics.

Lastly, another feature that researchers can utilize is human posture prediction from frames taken by event-based cameras. Human pose estimations can give more information to the researcher about the person's movements \cite{10208530}. There have been lots of research understanding and detecting human positions from events from frames taken on event-based cameras \cite{10208530, 10678345, 9157340, 9522759}. Being able to understand the human poses that are captured in the events can allow researchers to understand or predict the actions of the people that are in cities, which are integral to the city dynamics themselves. This can allow researchers to understand whether people are sitting, standing, slouching, running, and more all to better understand the activities of the city's inhabitants.

Through the use of neural networks \cite{8660483}, clustering techniques \cite{9607843}, and other machine learning and computer vision techniques, researchers are able to utilize event-based cameras to gather large amounts of information in an efficient way. Although event-based cameras have the challenge that they provide asynchronous data \cite{10517639}, technological advancements have overcome this challenge and now lots of information can be gathered from event-based cameras like optical flow, object detection and tracking, and human pose estimations.

\section{Multi-Sensor Fusion}
While event-based cameras already offer strong advantages in low-light performance, low-latency motion capture, and privacy-preserving imaging, several avenues remain open for expanding their usefulness in urban dynamics research, especially when combining with other sensing modalities.

\subsection{Event-based Visible and Infrared Fusion}
One promising direction is the fusion of event-based cameras (EBCs) with infrared (IR) sensors. IR cameras excel at detecting heat signatures and are widely used for nighttime monitoring, wildlife tracking, and safety applications \cite{mccafferty2007value, sarawade2018infrared}. A recent study has begun exploring IR + event-based fusion \cite{Geng_2024_CVPR}, demonstrating that combining the asynchronous motion sensitivity of EBCs with the thermal information from IR imaging can overcome limitations of each modality individually.

Event cameras capture motion exceptionally well, boasting an extremely low latency (on the order of µs) and a very high dynamic range (typically > 120 dB), which allow them to overcome motion blur and remain robust under a wide range of lighting conditions. However, their output is a stream of asynchronous and spatially sparse events, fundamentally different from traditional dense image data \cite{Gallego_2022}. In contrast, infrared (IR) sensors capture thermal information that is not affected by ambient light and provide consistent thermal structure (e.g., human silhouettes), but they are not immune to motion blur, particularly uncooled microbolometers, which blur due to the thermal inertia of each pixel \cite{ramanagopal2020pixelwisemotiondeblurringthermal}.

The Event-based Visible and Infrared Fusion (EVIF) system proposed by Geng et al. \cite{Geng_2024_CVPR} shows how event data can be used to improve both visible-image reconstruction and infrared deblurring. Because event cameras offer extremely high temporal resolution and a very wide dynamic range, they can capture sharp motion information even in scenes with fast movement, strong glare, or severe lighting changes—conditions where standard RGB cameras easily blur or over/underexpose \cite{8946715, 9729634}. EVIF takes advantage of this by using a cross-task event enhancement module that passes detailed event-based features to the IR deblurring network, helping it recover sharper thermal images, and by applying a bi-level min–max mutual information optimization strategy that reduces redundant information while preserving important complementary cues from both sensors \cite{Geng_2024_CVPR}.

For urban dynamics research, event–IR fusion offers several important advantages. City environments often involve rapid motion, challenging illumination, and nighttime activity where RGB cameras struggle \cite{9729634, chen2018learningdark}. By combining the sharp motion cues from event cameras with the stable thermal patterns captured by IR sensors, EVIF-style systems can improve tasks such as estimating crowd movement, detecting unusual behavior, and tracking multiple people or vehicles simultaneously. Moreover, because event cameras capture only sparse changes in brightness rather than full textures or identifiable facial details, event–IR fusion can support these analytics while maintaining a higher degree of privacy compared to traditional RGB-based monitoring \cite{Geng_2024_CVPR}.

\subsection{Event–LiDAR Fusion for Robust 3D Perception}
A second direction is the fusion of event-based cameras with LiDAR to improve 3D perception in fast-changing urban scenes. This fusion addresses the distinct drawbacks of each sensor by leveraging LiDAR’s strong geometric accuracy and the high-frequency motion sensing of event cameras \cite{s22249577}.

LiDAR provides accurate depth and reliable 3-D structural information but typically operates at low frame rates and can distort fast-moving objects during a scan \cite{behley2018efficient}. Moreover, LiDAR measurements often degrade severely in low-visibility conditions (e.g., rain, fog) due to light scattering \cite{dreissig2023surveylidarperceptionadverse}. For tasks such as simultaneous localization and mapping (SLAM) or odometry, LiDAR-based systems are also prone to drift in geometrically homogeneous environments (e.g., long corridors) where structural features are scarce \cite{10924516, Huang2021ReviewOL}.

Event cameras can partially compensate for these limitations by providing high-frequency motion cues that improve tracking during fast dynamics and by contributing additional temporal structure that helps reduce drift in feature-poor environments \cite{rebecq2016evo}. Although event cameras are also affected by adverse weather, they can remain functional under certain low-visibility conditions when appropriately tuned \cite{11074301}.

This combination mirrors the benefits seen in event–IR fusion: LiDAR supplies stable geometric structure, while events add high-frequency motion information. Together, they can refine the trajectories of pedestrians, cyclists, and vehicles, reduce motion distortion, and provide smoother temporal updates.

For urban dynamics research, combining sparse events with LiDAR depth enables detailed movement analysis while preserving privacy, since neither modality captures identifiable textures. Potential applications include multi-target tracking, anomaly detection, and real-time flow estimation in dense city spaces.

\subsection{Event–Vibration Fusion for Improved Crowd Sensing}
Floor-vibration sensing has recently become an effective way to monitor large groups of people in public spaces. Systems such as GameVibes \cite{10.1145/3600100.3623750} show that vibration signals captured from the floor can reveal both crowd reactions (e.g., clapping, stomping, coordinated cheering) and crowd traffic (e.g., how many people enter or exit through each door). This approach is cost-efficient and privacy-friendly, but vibration signals alone can be uncertain because thousands of people can generate overlapping and highly variable patterns.

The combination of event-based cameras with vibration sensing can address many of these challenges. Event cameras provide high-frequency, low-latency motion information that captures fast or subtle changes in crowd activity. These rapid cues can help clarify ambiguous vibration signals during moments when many people move at once and the vibration patterns become difficult to separate. In this way, event data can help identify when reactions start and stop, how intense they are, or whether different motions overlap.

By combining these two sensing modalities, event–vibration fusion can provide a more complete understanding of crowd behavior. Event cameras contribute precise timing and motion details, while vibration sensing provides large-scale, structure-level information that reflects how the entire crowd behaves. Together, they can support more robust, privacy-preserving crowd analysis for tasks such as reaction detection, crowd-flow monitoring, and real-time safety assessment in large venues.

\section{Future Directions for Event-Based Sensing in Urban Dynamics}
Event-based sensing has strong potential for understanding how people move and interact in cities, but several areas still need further research.

Future work should evaluate event-based systems in real city environments such as sidewalks, transit stations, parks, and public venues. Testing at this scale will help us understand how event data changes across seasons, crowd levels, and different types of urban activity.

Urban dynamics research will also benefit from richer models of movement and interaction. Combining event data with complementary sensors like infrared, LiDAR, or vibration sensors can provide a more complete view of crowd flows, congestion, and rare or unusual events.

Because collecting labels in cities is difficult, future methods should focus on learning from unlabeled or weakly labeled event streams. Approaches that can discover patterns in crowd movement, trajectories, or activity categories without heavy annotation will be essential for scalable monitoring.

Privacy must remain a core priority. Since urban research often takes place in public spaces, event-based systems should produce analyses that protect individuals while still revealing meaningful aggregate patterns.

Finally, event data needs to be translated into tools that urban researchers and planners can use. Real-time flow maps, density trends, and indicators of unusual behavior will help connect event-based sensing with practical decisions about mobility, safety, and city design.

\section{Conclusion}
In this survey paper, we explored the topic of urban dynamic
research and how technology has shaped its data collection,
findings, and impact. We reviewed early work based on manual observation and traditional cameras and highlighted the limitations these methods face in scalability, efficiency, and privacy. We then explored how event-based cameras offer a fundamentally different approach to capturing motion and activity, providing high temporal resolution, low power consumption, and reduced identifiable detail. These characteristics position event-based sensing as a promising tool for studying how people move and interact in complex urban environments.

We also discussed the role of machine learning in interpreting event streams and the value of combining event-based cameras with other sensing modalities to obtain richer and more reliable information about city dynamics. Although event-based cameras still face challenges such as noise, data representation, and practical deployment, ongoing progress in algorithms, sensor fusion, and privacy-conscious design is steadily improving their capabilities.

Event-based cameras provide a promising foundation for more adaptable and privacy-aware urban sensing. As the technology improves, it may offer researchers clearer insight into how people move through cities and support better decisions for shaping urban life.

\bibliographystyle{ACM-Reference-Format}
\bibliography{references}

\end{document}